\UseRawInputEncoding
\documentclass{article}
\usepackage{iclr2027_conference,times}

\iclrfinalcopy

\usepackage{pifont}
\usepackage{multirow}
\usepackage{microtype}
\usepackage{url}
\usepackage{longtable}
\usepackage{booktabs}
\usepackage[utf8]{inputenc}
\usepackage{graphicx}
\usepackage{amsmath, amssymb, amsthm}
\usepackage{amsfonts}
\usepackage{caption}
\usepackage{subcaption}
\usepackage{float}
\usepackage{enumitem}
\usepackage{algorithm}
\usepackage{algpseudocode}
\usepackage{hyperref}
\usepackage[table]{xcolor}
\usepackage{colortbl}
\usepackage{makecell}
\usepackage{wrapfig}
\usepackage{placeins}
\usepackage{newunicodechar}
\usepackage{array}
\usepackage{booktabs}

\usepackage{tikz}
\usetikzlibrary{arrows.meta,positioning,fit,backgrounds,calc, shapes.geometric,decorations.pathreplacing}

\usepackage{pgfplots}
\pgfplotsset{compat=1.18}
\usepgfplotslibrary{fillbetween}

\usepgfplotslibrary{groupplots}

\usepackage[T1]{fontenc}
\usepackage{textcomp}

\usepackage{listings}

\usepackage{siunitx}
\newunicodechar{â}{-}
\definecolor{darkblue}{rgb}{0,0,0.5}
\definecolor{highlightgreen}{RGB}{144,238,144}
\definecolor{highlightblue}{RGB}{230,240,250}
\hypersetup{colorlinks=true, citecolor=darkblue, linkcolor=darkblue, urlcolor=darkblue}

\lstdefinestyle{prompt}{
  basicstyle=\ttfamily\scriptsize,
  columns=fullflexible,
  keepspaces=true,
  breaklines=true,
  breakatwhitespace=true,
  breakindent=0pt,
  showstringspaces=false,
  frame=leftline,
  framerule=0.8pt,
  rulecolor=\color{black!25},
  framesep=6pt,
  xleftmargin=8pt,
  aboveskip=0.5\baselineskip,
  belowskip=0.9\baselineskip,
  upquote=true,
  literate={--}{{-{}-}}2
}
 
\definecolor{headerbg}{RGB}{221,235,247}   % soft blue header row
\definecolor{oursbg}{RGB}{255,246,214}      % soft gold for our rows/cols
\definecolor{bestbg}{RGB}{201,231,192}      % soft green for best value
\definecolor{grouprowbg}{RGB}{242,242,242}  % light grey group rows
\definecolor{rulegrey}{RGB}{130,130,130}
\newcommand{\hdrrow}{\rowcolor{headerbg}}
\newcommand{\grouprow}{\rowcolor{grouprowbg}}
\newcommand{\oursrow}{\rowcolor{oursbg}}
\newcolumntype{O}{>{\columncolor{oursbg}}r}
\usepackage{tcolorbox}
\tcbuselibrary{skins,breakable}
\definecolor{keybg}{RGB}{234,242,250}
\definecolor{keybar}{RGB}{31,78,121}
\newtcolorbox{keybox}{enhanced, breakable, boxrule=0pt, frame hidden,
  colback=keybg, borderline west={3pt}{0pt}{keybar}, arc=1pt,
  left=8pt, right=8pt, top=3pt, bottom=3pt, boxsep=0pt,
  before skip=4pt, after skip=4pt}

\usepackage{xcolor}

\algrenewcommand\algorithmicindent{1.0em}

\title{Compiling Expert Procedures into Verified Programs: \\ When a Small Model Is Enough}

\author{%
Vishvesh Bhat\textsuperscript{1}\qquad
Asad Aali\textsuperscript{2}\qquad
Muhammad Ahmed Mohsin\textsuperscript{2}\\
\textsuperscript{1}CoreThink AI\qquad
\textsuperscript{2}Stanford University
}

\begin{document}

\date{}
\maketitle

\begin{abstract}
In procedural tool-use domains, how much of the gap between base models is
procedure rather than capability? We build an explicit program from a
teacher's traces and run it in front of unchanged small models, including
an 8B model. The construction, \textsc{PLVR}, takes the reasoning traces
of a teacher, Kimi-K3, clusters their segments to locate recurring operations, has a frontier
model author one typed primitive per operation, each run by a small frozen
model and checked by a deterministic verifier, and assembles the program by
A$^*$ search over a rule library mined from the same traces, under a
reference-action objective: the distance between the actions the program
takes and the actions the teacher took on the training tasks. At inference
the program takes one of two forms: for troubleshooting and tool search it
produces the chain of thought the student would otherwise generate for
itself, and the student acts on it; for multi-turn tool matching the
student proposes each action and the program checks, repairs, or retries it
before it goes out. On the official $\tau^2$-Bench Retail and Telecom test
sets and on LiveMCPBench, with three Nemotron base models and the same
program for each, the answer depends on the domain. On Telecom,
Granite-4.2-8B as the student goes from 43.0\% to 91.2\% under the
program, and every Nemotron student reaches 93.3--94.4\%, against at most 58.2\% for
fine-tuning on the same traces, prompt optimisation of the student, or two
token-matched baselines. With Granite-4.2-8B as the student, the program
is within 2 to 3 points of the Nemotron students on Telecom and within 3
on Retail, and 3 to 13 points below them on LiveMCPBench: on $\tau^2$
Telecom an 8B student is enough, and on tool search the student still
matters. The Telecom result is in-domain compositional generalisation: the
program is built from synthetic tasks in the benchmark's own environment
and run on the benchmark's unseen test tasks. On Retail and LiveMCPBench
the program is competitive with the best baseline. The spread between base models shrinks from about 7
points to about 2, averaged over the suites, with no weights changed.
\end{abstract}

%% ============================================================
%% SECTION 1: INTRODUCTION
%% ============================================================

\section{Introduction}

A small model that fails a multi-step tool-use task may lack the capability
for the task, or may only lack the procedure: which record to read first,
which check to run before which fix, when to stop. The two look alike in an
end-to-end score. They differ in what fixes them. Capability is addressed by
a bigger or better-trained model; procedure can be written down.

This paper asks how much of the gap between base models in procedural
tool-use domains is procedure. We build an explicit program from a
teacher's traces and run it in front of unchanged models of different
sizes, down to an 8B model. If the gap is mostly procedure, the choice of
model under the program should matter little; if it is mostly capability,
the program should help the stronger model more and the model should still
matter. The answer we find depends on the domain: on troubleshooting and
multi-turn tool matching an 8B student under the program is level with
larger ones, and on tool search the student still matters.

\paragraph{Trace-guided program construction.} \textsc{PLVR} takes the teacher traces
that supervised fine-tuning would consume and decompose them instead.
Clustering the traces' segments proposes where recurring operations occur;
a frontier model decides what each operation is and writes it as a
\emph{primitive}, a call to a small frozen model scoped to one operation,
with a typed input, an output schema, and a deterministic verifier that
rejects ill-formed or ungrounded outputs; and the primitives are composed
into a program by search. The student's weights are never updated; the
program is the object that is built.

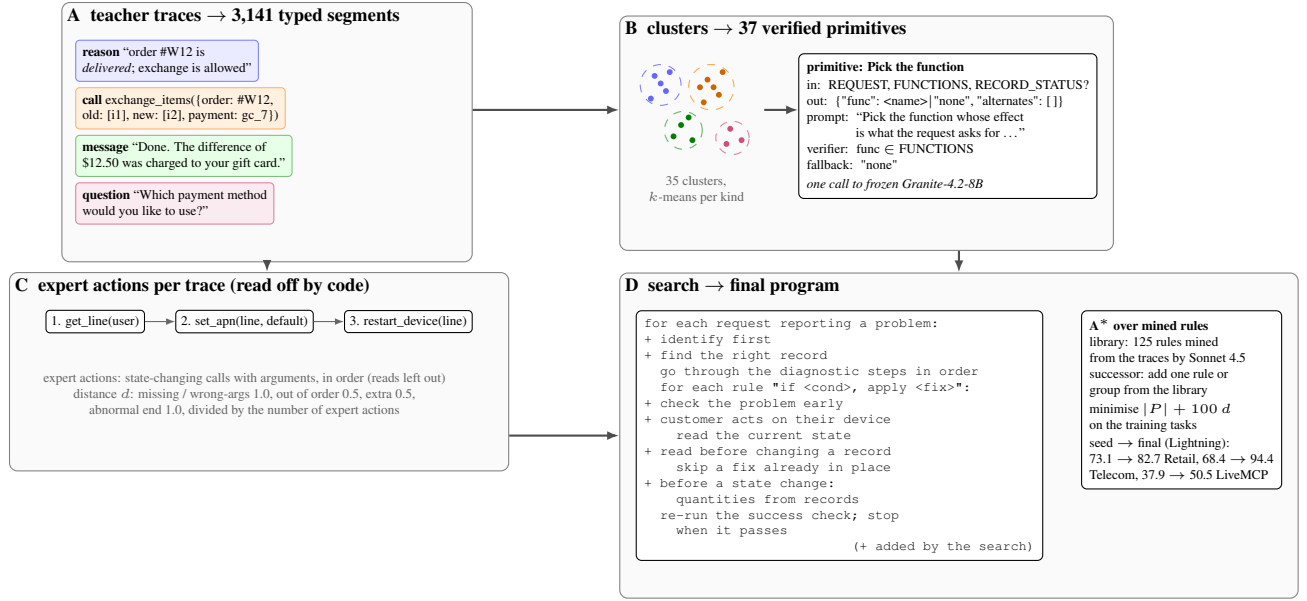
\begin{figure}[t]
\centering
\begin{tikzpicture}[
  font=\scriptsize,
  panel/.style={draw=black!50, rounded corners=3pt, inner sep=5pt, fill=black!2},
  ptitle/.style={font=\scriptsize\bfseries, anchor=north west, inner sep=0pt},
  seg/.style={draw, rounded corners=1.5pt, inner sep=2.5pt, font=\tiny, align=left},
  reason/.style={seg, fill=blue!8, draw=blue!50},
  call/.style={seg, fill=orange!12, draw=orange!60},
  msg/.style={seg, fill=green!10, draw=green!50!black!50},
  qst/.style={seg, fill=purple!8, draw=purple!50},
  card/.style={draw, rounded corners=2pt, fill=white, inner sep=3pt, font=\tiny, align=left},
  pnode/.style={draw, rounded corners=1.5pt, fill=white, inner sep=2pt, font=\tiny},
  arr/.style={-{Latex[length=1.6mm]}, thick, black!70},
  tarr/.style={-{Latex[length=1.2mm]}, thin, black!60},
  note/.style={font=\tiny, text=black!60, align=center}
]
%% ---------------- Panel A: traces -> segments
\node[reason] (s1) at (0,0) {\textbf{reason} ``order \#W12 is\\\textit{delivered}; exchange is allowed''};
\node[call, below=2pt of s1.south west, anchor=north west] (s2) {\textbf{call} exchange\_items(\{order: \#W12,\\ old: [i1], new: [i2], payment: gc\_7\})};
\node[msg, below=2pt of s2.south west, anchor=north west] (s3) {\textbf{message} ``Done. The difference of\\\$12.50 was charged to your gift card.''};
\node[qst, below=2pt of s3.south west, anchor=north west] (s4) {\textbf{question} ``Which payment method\\would you like to use?''};
\begin{scope}[on background layer]
\coordinate (aR) at ([xshift=22mm]s3.east);
\node[panel, fit=(s1)(s2)(s3)(s4)(aR), inner ysep=14pt, inner xsep=5pt] (pA) {};
\end{scope}
\node[ptitle] at ([xshift=2pt,yshift=-2pt]pA.north west) {A\; teacher traces $\to$ 3,141 typed segments};

%% ---------------- Panel B: clusters -> primitives
\node[card, anchor=north west] (card) at (8.35,0.1) {%
\textbf{primitive: Pick the function}\\[1pt]
in:\; REQUEST, FUNCTIONS, RECORD\_STATUS?\\
out:\; \{"func": <name>$|$"none", "alternates": [\,]\}\\
prompt:\; ``Pick the function whose effect\\ \phantom{prompt:\;} is what the request asks for \ldots''\\
verifier:\; func $\in$ FUNCTIONS\\
fallback:\; "none"\\[1pt]
\textit{one call to frozen Granite-4.2-8B}};
\begin{scope}[shift={(6.35,-0.1)}]
  \foreach \x/\y in {0.1/-0.1,0.25/-0.3,0.05/-0.4,0.3/-0.05,0.18/-0.2} \fill[blue!60] (\x,\y) circle (1.1pt);
  \foreach \x/\y in {0.8/-0.15,0.95/-0.35,0.75/-0.45,1.0/-0.05,0.88/-0.25,0.7/-0.25} \fill[orange!80!black] (\x,\y) circle (1.1pt);
  \foreach \x/\y in {0.45/-0.75,0.6/-0.95,0.35/-0.9,0.55/-0.65} \fill[green!50!black] (\x,\y) circle (1.1pt);
  \foreach \x/\y in {1.1/-0.8,1.25/-0.95,1.05/-1.0} \fill[purple!70] (\x,\y) circle (1.1pt);
  \draw[blue!50, dashed] (0.18,-0.22) circle (0.27);
  \draw[orange!70, dashed] (0.85,-0.25) circle (0.3);
  \draw[green!50!black!60, dashed] (0.48,-0.82) circle (0.25);
  \draw[purple!50, dashed] (1.13,-0.92) circle (0.22);
  \node[note, anchor=north] (scnote) at (0.65,-1.3) {35 clusters,\\$k$-means per kind};
  \coordinate (scTL) at (-0.15,0.15);
  \coordinate (scR) at (1.55,-0.55);
\end{scope}
\draw[arr] (scR) -- (card.west |- scR);
\begin{scope}[on background layer]
\node[panel, fit=(scTL)(scnote)(card), inner ysep=14pt, inner xsep=6pt] (pB) {};
\end{scope}
\node[ptitle] at ([xshift=2pt,yshift=-2pt]pB.north west) {B\; clusters $\to$ 37 verified primitives};
\draw[arr] (pA.east |- scR) -- (pB.west |- scR);

%% ---------------- Panel C: expert actions per trace
\begin{scope}[shift={(-0.95,-3.45)}]
\node[pnode] (n1) at (0,0) {1. get\_line(user)};
\node[pnode, right=4mm of n1] (n4) {2. set\_apn(line, default)};
\node[pnode, right=4mm of n4] (n7) {3. restart\_device(line)};
\draw[tarr] (n1) -- (n4); \draw[tarr] (n4) -- (n7);
\node[note, below=4mm of n4] (cnote) {expert actions: state-changing calls with arguments, in order (reads left out)\\distance $d$: missing / wrong-args 1.0, out of order 0.5, extra 0.5,\\abnormal end 1.0, divided by the number of expert actions};
\coordinate (cS) at ([yshift=-1mm]cnote.south);
\coordinate (cW) at ([xshift=-3mm]n1.west);
\coordinate (cE) at ([xshift=3mm]n7.east);
\end{scope}
\begin{scope}[on background layer]
\node[panel, fit=(n1)(n4)(n7)(cS)(cW)(cE)(cnote), inner ysep=14pt] (pC) {};
\end{scope}
\node[ptitle] at ([xshift=2pt,yshift=-2pt]pC.north west) {C\; expert actions per trace (read off by code)};
\draw[arr] (pA.south) -- (pA.south |- pC.north);

%% ---------------- Panel D: search -> program
\node[card, font=\tiny\ttfamily, anchor=north west, align=left] (prog) at (6.2,-3.3) {%
for each request reporting a problem:\\
+ identify first\\
+ find the right record\\
\ \ go through the diagnostic steps in order\\
\ \ for each rule "if <cond>, apply <fix>":\\
+   check the problem early\\
+   customer acts on their device\\
\ \ \ \ read the current state\\
+   read before changing a record\\
\ \ \ \ skip a fix already in place\\
+ before a state change:\\
\ \ \ \ quantities from records\\
\ \ re-run the success check; stop\\
\ \ \ \ when it passes\\
\ \ \ \ \ \ \ \ \ \ \ \ \ \ \ \ \ \ \ \ \ \ \ \ \ \ (+ added by the search)};
\node[card, right=5mm of prog.north east, anchor=north west, align=left] (astar) {%
\textbf{A$^*$ over mined rules}\\
library: 125 rules mined\\ from the traces by Sonnet 4.5\\
successor: add one rule or\\ group from the library\\[1pt]
minimise $|P| + 100\,d$\\ on the training tasks\\[1pt]
seed $\to$ final (Lightning):\\
73.1 $\to$ 82.7 Retail, 68.4 $\to$ 94.4\\ Telecom, 37.9 $\to$ 50.5 LiveMCP};
\begin{scope}[on background layer]
\node[panel, fit=(prog)(astar), inner ysep=14pt, inner xsep=6pt] (pD) {};
\end{scope}
\node[ptitle] at ([xshift=2pt,yshift=-2pt]pD.north west) {D\; search $\to$ final program};
\draw[arr] (pB.south -| pD.north) -- (pD.north);
\draw[arr] (pC.east |- prog.west) -- (pD.west |- prog.west);
\end{tikzpicture}
\caption{Trace-guided program construction. (A) Each kept teacher trace is cut into
segments typed as reason, call, message, or question. (B) Segments are
clustered within kind; each cluster's behaviours become primitives, each a
single call to frozen Granite-4.2-8B with a typed input, an output schema, a
deterministic verifier, and a fallback (\S\ref{sec:primitives}). (C) From
the same traces, code alone reads off the expert actions, and a program is
scored by the distance between the actions it takes and the expert's
(\S\ref{sec:process}). (D) A$^*$ adds rules from a library mined from the traces, minimising $f = |P| + 100\,d$ on the training tasks, and
keeps the best program (\S\ref{sec:search}).}
\label{fig:pipeline}
\end{figure}

\paragraph{Two forms of the program.}\label{sec:intro-forms} A base model asked to reason before
acting generates a plan for itself at inference time and then follows it.
For troubleshooting and tool search, the program is that plan, constructed once,
from the teacher's traces, as a typed composition of primitives, and at
inference runs it to produce the chain of thought the student then acts on:
the program replaces the student's reasoning. For multi-turn tool matching
the program is a guard around the student: the student proposes each action,
and the program checks it against the records, the rules, and the user's
consent, repairs it, or sends it back to be proposed again. In both forms
the student is the same model with the same weights; what changes is whether
its reasoning is written for it or its actions are checked for it.

\paragraph{Finding the program.} Assembling the program is a discrete
search with no gradients. A frontier model first mines a library of rules
from the traces, one line each. Starting from a hand-written seed, A$^*$
adds rules from that library, scoring each partial program by the distance
between the actions
a program takes and the actions the teacher took on the same task, read off
the trace by code: a reference-action objective. We report the seed's and
the final program's accuracy on the test suites so the search's
contribution is visible.

\paragraph{Results.} Four results, all in Table~\ref{tab:main}. First, on
$\tau^2$ Telecom, Granite-4.2-8B as the student goes from 43.0\% to 91.2\% under the
program, and Nemotron 3.5 Lightning, Nemotron 3 Nano, and
Nemotron 3 Super reach 94.4\%, 93.3\%, and 93.3\%; no other arm on any
student exceeds 58.2\%. Second, with Granite-4.2-8B as the student the
program is 2.1 to 3.2 points below the Nemotron students on Telecom, within
2.7 points of them on Retail, and 3.2 to 12.6 points below on LiveMCPBench:
on $\tau^2$ Telecom an 8B student is enough, and on tool search the
student still matters. On Retail and LiveMCPBench the program is
competitive with the best baseline, above it on some students and below it
on others. Third, the spread between the three students shrinks from 6.6
points to 1.9 averaged over the suites, with no weights changed; per suite,
from 15.8 to 3.5 on Retail, 15.0 to 1.1 on Telecom, and 10.9 to 9.4 on
LiveMCPBench. Fourth, the search contributes: on Lightning it takes the
hand-written seed from 73.1\%, 68.4\%, and 37.9\% on the three suites to
82.7\%, 94.4\%, and 50.5\% (59.8\% to 75.9\% averaged).

\paragraph{Scope.} The method needs intermediate outputs that admit checking.
Tasks whose quality is a matter of judgement rather than satisfaction, such
as summarisation and open generation, are outside its scope, and we make no
claim about them. Extraction was run on one teacher and three kinds of
tool-use task, and we do not
claim the primitives transfer to other domains.

\paragraph{Contributions.}
\begin{itemize}
  \item \textbf{An estimate of how much of the gap is procedure.} One
    program, the same for every base model, run in front of three Nemotron
    students and of Granite-4.2-8B on three suites. With the 8B model as the
    student the program reaches 91.2\% on Telecom and 82.5\% on Retail,
    within 3 points of the Nemotron students; on LiveMCPBench the Nemotron
    students are 3 to 13 points ahead of it (\S\ref{sec:results}).
  \item \textbf{Trace-guided program construction (\textsc{PLVR}).} A frontier model
    authors 37 primitives from clusters of teacher-trace segments, each a
    typed call to a small frozen model carrying a deterministic verifier,
    so that every primitive is checked as it runs; a rule library is mined
    from the same traces; and A$^*$ assembles the program under a
    reference-action objective read off the traces by code
    (\S\ref{sec:method}, Appendix~\ref{app:extraction}). The search itself
    is plain A$^*$ and is not claimed as a contribution.
\end{itemize}

%% ============================================================
%% SECTION 2: RELATED WORK
%% ============================================================

\section{Related Work}
\label{sec:related}

\paragraph{Distillation and post-training.} Knowledge distillation transfers a
teacher's behaviour into a student's parameters, classically through soft
targets and in the LLM setting through supervised fine-tuning on the teacher's
outputs or reasoning traces \citep{instructgpt}. RLHF fits reward models to
human preferences \citep{rlhf,instructgpt}; RLVR instead rewards
programmatic checks such as test execution or answer matching
\citep{guo2025deepseekr1}. \textsc{PLVR} consumes the same traces as
trace-based SFT, but puts the result into a program over frozen components
rather than into weights, and the student's parameters are not changed; its
objective is the distance to the teacher's actions rather than a reward.

\paragraph{Reasoning as programs.} Program-of-thought and PAL-style
methods have a model emit executable code in place of free-text reasoning
\citep{pal,programofthoughts}, and later work distils such programs into
smaller models \citep{pad}. These write the program to a fixed template
per example. \textsc{PLVR} writes one program per task family, from operations
mined out of the teacher's traces, and runs it either to produce the
reasoning the student then follows or, in its guard form, to check and
repair the actions the student proposes; the second form is closer to
runtime guardrails and output verifiers than to program-of-thought.

\paragraph{Process supervision and compound-system optimisation.} Process
supervision scores intermediate steps \citep{uesato2022process,
lightman2023verify, wang2024mathshepherd, agentprm2026, kamoi2025formal}.
ProTeGi, TextGrad, Trace, and GEPA optimise language-model systems through
textual or trace feedback without derivatives through the model calls
\citep{protegi,textgrad,cheng2024trace,agrawal2026gepa}. DSPy,
ADAS, AFlow, and AgentFlow compile or search declarative pipelines and
workflows \citep{dspy,hu2025adas,aflow,agentflow}. Logic-LM, Proof-of-Thought, and SEVerA couple language models
with symbolic structure and executable checks \citep{logiclm,pot,severa}. \textsc{PLVR} differs from these in where
its components come from: the primitives are authored from clusters of a
teacher's traces rather than declared in advance, and the process signal is read off
those same traces rather than learned or annotated. The search itself is plain A$^*$ over a mined rule library and is not
claimed as a contribution.

%% ============================================================
%% SECTION 3: METHOD
%% ============================================================

\section{Methodology}
\label{sec:method}

A \textsc{PLVR} program is a composition of typed primitives arranged by a
small set of operators. We describe the training data and teacher traces
(\S\ref{sec:data}), the primitive library (\S\ref{sec:primitives}), the
expert actions read off each trace and the distance to them
(\S\ref{sec:process}), and the search (\S\ref{sec:search}); the primitives
and the clusters each is drawn from are in Appendix~\ref{app:extraction}.

\subsection{Training data and teacher traces}
\label{sec:data}

The three evaluation suites differ in the shape of their tasks, so we
built a synthetic training set of 553 tasks in three subsets, one per kind
of task (Table~\ref{tab:trainset}). The \emph{retail}
subset is 142 tasks in the $\tau^2$ retail format, multi-turn tool matching
against a policy. The \emph{telecom} subset is 178 tasks in the $\tau^2$
telecom format, mostly troubleshooting. The \emph{tool-search} subset is 233
tasks in the LiveMCPBench format, tool search over a large registry. Each
subset is split into training and held-aside tasks: 119/23 for retail,
122/56 for telecom, 183/50 for tool-search.

\paragraph{Generation.} Claude 5 Opus generated the tasks. For retail and
telecom it used $\tau^2$-Bench's own tools, environment, and policies and
wrote new questions against them; for tool-search it used LiveMCPBench's
registry format and wrote new questions against that.

\paragraph{Overlap guard.} No benchmark task, and nothing that could stand
in for one, appears in the training set:
\begin{enumerate}[nosep,leftmargin=*]
  \item \emph{No benchmark instances.} Every benchmark task is excluded by
    id, including held-out ones. On LiveMCPBench the benchmark's tasks were
    never read at all.
  \item \emph{No shared underlying data.} A training task may not reuse
    what a benchmark task is built on: the same issue type and fault set
    (telecom), or the same customer or order (retail).
  \item \emph{No near-duplicates.} Training task text is checked
    mechanically against benchmark text, and close matches are rewritten
    or dropped.
  \item \emph{Held-aside tasks stay out.} Each subset's held-aside split
    is fixed up front and never used for training or search.
  \item \emph{No uninformative tasks.} Tasks that always pass or cannot be
    solved are removed.
\end{enumerate}

Kimi-K3 is run on every task with the task's tools and a system prompt that
asks it to work through the task, record its steps, and finish with the tool
calls; it names no operation and no primitive. We keep the reasoning the
provider returns separately from the reply. Traces that do not reach the
task's ground truth are discarded: retail and telecom traces must re-score
1.0 under the $\tau^2$ evaluator, and tool-search traces must pass both an
LLM judge and the task's deterministic label. Every telecom and tool-search
trace is kept; on retail, 128 of 142 are (90.1\%). The 539 kept traces are
the only demonstrations in this paper: they are what the primitives are
extracted from, what the search scores against, and what every baseline is
trained on. Two frontier models take part in building the program and see
no benchmark data: Claude 5 Opus writes the primitives from the clustered
traces (\S\ref{sec:primitives}), and Claude Sonnet 4.5 mines the rule
library the program search draws on (\S\ref{sec:search}).

\begin{table}[h]
\centering
\small
\setlength{\tabcolsep}{5pt}
\begin{tabular}{@{}llrrrr@{}}
\toprule
\hdrrow Subset & Format & Tasks & Train & Held aside & Kept traces \\
\midrule
Retail      & $\tau^2$ retail format, multi-turn tool matching & 142 & 119 & 23 & 128 \\
Telecom     & $\tau^2$ telecom format, troubleshooting & 178 & 122 & 56 & 178 \\
Tool-search & LiveMCPBench format, tool search & 233 & 183 & 50 & 233 \\
\midrule
Total       &  & 553 & 424 & 129 & 539 \\
\bottomrule
\end{tabular}
\caption{The training set and the kept teacher traces.}
\label{tab:trainset}
\end{table}

\subsection{Primitive library}
\label{sec:primitives}

The primitives are trace-grounded and model-authored. The kept traces are
processed by segmenting each trace, typing each segment as a \emph{reason}, a
\emph{call}, a \emph{message}, or a \emph{question}, embedding, and
clustering with $k$-means within each kind. This gives 3,141 segments in 35
clusters: 15 reason, 8 call, 9 message, 3 question (segmentation rule and
embedding model in Appendix~\ref{app:extraction}). Clustering locates where
an operation lives in the traces; it does not by itself separate which
decision a segment witnesses, since one segment can show several (a call
segment shows the function chosen and the parameters filled). Clustering
therefore proposes where operations occur; a frontier model, Claude 5 Opus,
post-processes the clusters, decides what the operations are, and writes
each primitive: its input--output signature, prompt, verifier, and fallback.
The result is 37 primitives in six groups: understanding the message (7),
finding the tools (5), building the call (8), checking the policy (7),
troubleshooting conditions (6), and finishing (4). Each primitive is anchored
to the clusters its segments come from; a primitive draws on 1 to 14
clusters and on 1 to 195 segments, and a cluster carries 1 to 14 primitives.
Fifteen of the 37 are used in more than one subset; the rest are specific
to one kind of task. The full mapping between primitives and
clusters is in Appendix~\ref{app:extraction}.

\paragraph{What a primitive is.} Each primitive is one call to
Granite-4.2-8B, frozen, under its own prompt. It has named inputs plus a
shared context (the facts gathered so far, the latest user message, the
agent's previous message, and where relevant the policy), a JSON output
schema, a deterministic \emph{verifier}, and a \emph{fallback} returned when
the verifier rejects. No prompt is tuned against a benchmark or an end-task
reward. The verifier makes each primitive checkable: an
output that violates its schema, quotes text not verbatim in its source,
cites a fact that does not exist, or gives an expression that does not
evaluate is rejected and the fallback is used.

\paragraph{What is automatic and what is not.} Segmentation, typing,
embedding, and clustering involve no language model and no hand-picked
segments; the clusters, and which segments belong to them, are decided by
the clustering alone. The frontier model decides which operations the
clusters show and drafts each primitive's signature, prompt, verifier, and
fallback from the clusters' members. Names are the authors'. The claim is
not that the primitives were recovered automatically; it is that each one is
anchored to segments of the teacher's traces, and that the whole library was
authored from the kept traces at negligible compute.

\subsection{Expert actions and the distance to them}
\label{sec:process}

Each kept trace gives, by code alone, the \emph{expert actions} for its
task: the sequence of state-changing actions the teacher took (a device
action by the user, or a write or tool call by the agent), each with its
arguments; reads are left out. A program is scored by running it on a task
and comparing the conversation's sequence of actions with the expert
actions for the same task. The distance counts 1.0 for each expert action
that is missing or has wrong arguments, 0.5 for each action out of order,
0.5 for each extra state-changing action, and 1.0 for an abnormal ending,
divided by the number of expert actions. Lower is better, and a program's
score is the mean distance over the search set. This is the
reference-action objective the search minimises; no model judges a
conversation.

\subsection{Program search}
\label{sec:search}

A program is the seed plus a set of rules, each a line in the same form as
the seed's, and its primitives. The rules come from a library that Claude
Sonnet 4.5 mines from the kept traces once, before the search: 61 rules
from the tool-search traces and 64 from the telecom traces, each a one-line
behaviour the expert shows (Appendix~\ref{app:mined}). The combined
library of 125 rules is the candidate set for every block; the retail block
has no library of its own and draws on the rules mined from the telecom and
tool-search traces.

A$^*$ starts from the seed. A successor inserts one rule, or one group of
rules, from the library into the current program; it is kept only if the
program still compiles and runs without errors. Partial programs are scored by the distance of
\S\ref{sec:process} plus a small cost per rule,
\[
f(P) \;=\; |P| \;+\; 100\,d(P),
\]
where $|P|$ is the number of rules in the program and $d(P)$ is the mean
distance to the expert actions over the search set, so of two programs at
the same distance the smaller wins. The final program is the one with the
best task outcome on the training tasks among those the search reached.

\paragraph{The run.} The search scored 32 programs for the telecom block,
87 for the retail block, and 9 for the tool-search block (the seed, six
group edits, their merge, and the final program). Settings are in
Appendix~\ref{app:hyperparams}.

\paragraph{Seed and final program.} The seed is three hand-written blocks,
one per subset of the training set, joined into one program. The telecom
block walks the policy's checklist in order, reads the system state before
each fix rule, does not repeat a fix already in place, and re-runs the check
until it passes. The tool-search block retrieves candidate tools for each
request, picks one, fills its parameters, calls it, and checks whether the
request is achieved. The retail block takes the base model's proposed action
each turn and checks it before it goes out: unknown tools and malformed ids
are rejected, a change must take every value from a record that was read or
from the customer's words, the customer must be identified and own the
record and the policy rule must hold, a change goes out only after the
customer's yes and is never made twice, and a claimed ``done'' without the
call is retried. The search added rules to all three blocks. Both programs are
listed in full in Appendix~\ref{app:program}. On Lightning, the seed scores
73.1\% on Retail, 68.4\% on Telecom, and 37.9\% on LiveMCPBench; the final
program 82.7\%, 94.4\%, and 50.5\%.

%% ============================================================
%% SECTION 4: EXPERIMENTS
%% ============================================================

\section{Experiments}
\label{sec:setup}

\subsection{Benchmarks}
\label{sec:benchmarks}

We train on the set of \S\ref{sec:data} and evaluate on the official
$\tau^2$-Bench Retail and Telecom test sets and on LiveMCPBench.
$\tau^2$-Bench is multi-turn dialogue against a stated policy, judged by the
benchmark's user simulator, where a small tool set makes correctness turn on
ordering and admissibility. LiveMCPBench is tool selection and invocation over
a large MCP registry, where correctness turns on choosing the right tool from
many and binding arguments that resolve. Each suite has a training subset
in its format (\S\ref{sec:data}).

\subsection{Models}
\label{sec:models}

The primitives run on Granite-4.2-8B, frozen, each under its own prompt. The students are Nemotron 3.5 Lightning, Nemotron
3 Nano, and Nemotron 3 Super. Each receives every treatment: vanilla,
fine-tuning on the traces, GEPA, the two token-matched baselines, and the
program.

\subsection{Baselines}
\label{sec:baselines}

Every arm consumes the same kept teacher traces and the same synthetic
training set; no arm uses any other demonstrations.

\paragraph{Fine-tuning on the traces (SFT).} The student is fine-tuned on the
traces in two variants. The \emph{process} arm trains on the teacher's
reasoning followed by its final calls, so the student learns to reproduce the
procedure before answering. The \emph{outcome} arm trains on input to final
calls only, with the reasoning removed. Learning rate, LoRA rank, and epoch
are swept per model and variant and selected by solved count on the
held-aside traces of each subset. One fine-tune per model is evaluated on the
benchmarks and is the fine-tuning comparator in Table~\ref{tab:main}.
Construction and sweep are in Appendix~\ref{app:distill}.

\paragraph{Prompt optimisation (GEPA).} The student's own system prompt is
optimised with GEPA~\citep{agrawal2026gepa} on the training set, and the
student is then run as a native tool-calling agent with no program. This tests
whether optimising the student's prompt alone reaches what the program
reaches. Construction, budget, and the optimised prompt are in
Appendix~\ref{app:gepa-backbone}.

\paragraph{Vanilla.} The student with chain-of-thought prompting and no
other treatment. The program replaces this chain of thought, or checks
the actions it leads to.

\paragraph{Token-matched baselines.} Two arms spend the same tokens per task
as the final program, with Granite-4.2-8B doing the extra work, so that the
program is compared against other uses of the same budget on the same small
model. \emph{ReAct} runs a plan--act--reflect loop in the style of ReAct
\citep{yao2023react}: each turn, Granite-4.2-8B plans, the student acts from
that plan placed where its own reasoning would go, and Granite critics
review the action, with up to two revisions; the loop runs under the
program's Granite token budget (Appendix~\ref{app:react}). \emph{Sample-and-judge} samples
five candidate responses and has Granite-4.2-8B, as a judge, select one,
so that the total tokens match the program's. The judge prompt is in
Appendix~\ref{app:judge}.

\subsection{Protocol}
\label{sec:protocol}

The program is built once from the training set (\S\ref{sec:search}) and
held fixed. The program is evaluated three times on $\tau^2$ Telecom and
once on every other suite; every other arm is evaluated once per cell.
Reported scores are means over those runs. Token counts per task are
recorded for every arm, with the student's and Granite's tokens separated.

%% ============================================================
%% SECTION 5: RESULTS
%% ============================================================

\section{Results}
\label{sec:results}

Table~\ref{tab:main} reports every arm on $\tau^2$-Bench Retail and Telecom and
LiveMCPBench, per suite. The program rows run the same program, the same
primitives on the same frozen Granite-4.2-8B, in front of each student, with
no weights changed anywhere. Every cell except the program's Telecom cells
is a single run, so differences of a few points between arms are not
resolved; we describe them as competitive rather than ranking them.

\begin{table}[!htbp]
\centering
\caption{Accuracy (\%) by benchmark (pass@1 per run on $\tau^2$-Bench):
Retail 114 tasks, Telecom 114, LiveMCPBench 95. The program's
Telecom cells are means over three runs; all other cells, including the
Granite-as-student row, are one run. The best arm per student and column is in
bold. Averages are arithmetic means of the displayed cells. The SFT row is the one fine-tune per model of \S\ref{sec:baselines},
trained on the kept teacher traces. GEPA rows optimise the student's own
prompt with no program. ReAct and sample-and-judge rows spend the program's
token budget on Granite-4.2-8B without the program (\S\ref{sec:baselines}).
\textsc{PLVR} rows run the final program of \S\ref{sec:search}.}
\label{tab:main}
\small
\setlength{\tabcolsep}{4pt}
\begin{tabular}{@{}lrrrr@{}}
\toprule
\hdrrow & \multicolumn{2}{c}{$\tau^2$-Bench} & & \\
\hdrrow \cmidrule(lr){2-3}
\hdrrow System & Retail & Telecom & LiveMCP & Avg \\
\midrule
\grouprow \multicolumn{5}{@{}l}{\emph{Vanilla (chain-of-thought prompting)}} \\
Nemotron 3.5 Lightning & 75.4 & 47.4 & 42.5 & 55.1 \\
Nemotron 3 Nano        & 64.0 & 36.8 & \textbf{45.3} & 48.7 \\
Nemotron 3 Super       & 79.8 & 51.8 & 34.4 & 55.3 \\
\midrule
\grouprow \multicolumn{5}{@{}l}{\emph{SFT on the kept teacher traces}} \\
Nemotron 3.5 Lightning & 73.2 & 48.0 & 44.5 & 55.2 \\
Nemotron 3 Nano        & 77.0 & 51.0 & 44.0 & 57.3 \\
Nemotron 3 Super       & \textbf{86.3} & 50.0 & 57.4 & 64.6 \\
\midrule
\grouprow \multicolumn{5}{@{}l}{\emph{Prompt optimisation (GEPA on the student's prompt, no program)}} \\
Nemotron 3.5 Lightning & 79.8 & 51.8 & 47.0 & 59.5 \\
Nemotron 3 Nano        & 72.3 & 54.8 & 45.0 & 57.4 \\
Nemotron 3 Super       & 80.7 & 58.2 & 41.0 & 60.0 \\
\midrule
\grouprow \multicolumn{5}{@{}l}{\emph{ReAct planner loop on Granite-4.2-8B, token-matched}} \\
Nemotron 3.5 Lightning & 71.9 & 32.5 & 17.4 & 40.6 \\
Nemotron 3 Nano        & 64.0 & 28.1 & 18.4 & 36.8 \\
Nemotron 3 Super       & 69.3 & 28.1 & 23.7 & 40.4 \\
\midrule
\grouprow \multicolumn{5}{@{}l}{\emph{Sample-and-judge, best of 5, Granite-4.2-8B as judge, token-matched}} \\
Nemotron 3.5 Lightning & 82.5 & 55.3 & 42.1 & 60.0 \\
Nemotron 3 Super       & 83.3 & 46.5 & \textbf{58.9} & 62.9 \\
\midrule
\grouprow \multicolumn{5}{@{}l}{\emph{\textsc{PLVR}, final program (after search; ours)}} \\
\oursrow Nemotron 3.5 Lightning & \textbf{82.7} & \textbf{94.4} & \textbf{50.5} & \textbf{75.9} \\
\oursrow Nemotron 3 Nano        & \textbf{83.3} & \textbf{93.3} & \textbf{45.3} & \textbf{74.0} \\
\oursrow Nemotron 3 Super       & 79.8 & \textbf{93.3} & 54.7 & \textbf{75.9} \\
\midrule
\grouprow \multicolumn{5}{@{}l}{\emph{\textsc{PLVR}, seed program (hand-written, before search)}} \\
Nemotron 3.5 Lightning & 73.1 & 68.4 & 37.9 & 59.8 \\
\midrule
\grouprow \multicolumn{5}{@{}l}{\emph{Granite-4.2-8B alone}} \\
Vanilla                & 77.2 & 43.0 & 28.4 & 49.5 \\
\textsc{PLVR}, Granite as the student & 82.5 & 91.2 & 42.1 & 71.9 \\
\bottomrule
\end{tabular}
\end{table}

\subsection{How much of the result is the program}
\label{sec:spectrum}

\paragraph{Granite as the student.} Run with Granite-4.2-8B as the
student, producing the final response itself (one run per suite), the final
program scores 82.5\% on Retail, 91.2\% on Telecom, and 42.1\% on
LiveMCPBench (Table~\ref{tab:main}). Granite untreated scores 77.2\%,
43.0\%, and 28.4\%, so the program raises it by 5.3, 48.2, and 13.7 points.
Against the Nemotron students under the same program, Granite is 2.1 to 3.2
points below on Telecom, within 2.7 points on Retail (above Super, below
Lightning and Nano), and 3.2 to 12.6 points below on LiveMCPBench. On
Telecom and Retail an 8B student under the program is level with the larger
ones, and on Telecom it is within 3.2 points of the best result in the
table; on LiveMCPBench the student still matters.

\paragraph{The base model matters less.} No part of the program is tuned to
a particular student: the primitives run on Granite-4.2-8B, the search was
run once, and the three program rows differ only in which model reads the
chain of thought or has its actions checked. Untreated, the three students
are spread over 15.8 points on Retail, 15.0 on Telecom, and 10.9 on
LiveMCPBench (6.6 averaged over the suites). Under the program the spread is
3.5, 1.1, and 9.4 (1.9 averaged). The spread closes on the two suites where
an 8B student is level with the larger ones and stays on the suite where
the student matters.

\subsection{Against fine-tuning on the traces}
\label{sec:vs-monolithic}

Fine-tuning on the kept traces moves each student by a few points per
suite; its largest single gain over the untreated student is 23.0 points
(Super, LiveMCPBench). The program, built from the same traces, is ahead of
the fine-tune on Telecom by 46.4, 42.3, and 43.3 points on Lightning, Nano,
and Super. On Retail it is ahead by 9.5 and 6.3 points for Lightning and
Nano and behind by 6.5 for Super (79.8\% against 86.3\%); on LiveMCPBench
ahead by 6.0 and 1.3 for Lightning and Nano and behind by 2.7 for Super
(54.7\% against 57.4\%). The program's largest single gain over the
untreated student is 56.5 points (Nano, Telecom).

\subsection{Against prompt optimisation of the student}
\label{sec:vs-gepa}

GEPA optimises the student's own prompt with no program. Its largest gains
over the untreated student are on Nano (Telecom 36.8\% to 54.8\%, Retail
64.0\% to 72.3\%) and on Super's Telecom and LiveMCPBench (51.8\% to
58.2\%, 34.4\% to 41.0\%). The program is ahead of GEPA on Telecom by 42.6,
38.5, and 35.1 points. On Retail it is ahead by 2.9 and 11.0 for Lightning
and Nano and behind by 0.9 for Super (79.8\% against 80.7\%); on
LiveMCPBench ahead by 3.5, 0.3, and 13.7.

\subsection{Against the same token budget without the program}
\label{sec:vs-budget}

The two token-matched arms spend what the program spends, on the same
Granite-4.2-8B, without the program. The ReAct planner loop is below the
untreated student on every suite and every student except Nano on Retail
(64.0\% in both): spending the program's budget on a plan--act--reflect
loop makes the student worse, not better. Sample-and-judge with five
candidates is the strongest baseline on Retail and on Super's LiveMCPBench:
on Retail it is 0.2 points below the program for Lightning (82.5\% against
82.7\%) and 3.5 above it for Super (83.3\% against 79.8\%); on LiveMCPBench
8.4 below it for Lightning and 4.2 above it for Super (58.9\% against
54.7\%); on Telecom 39.1 and 46.8 points below it. Against the best
baseline in each cell, the program on Retail is within 0.2 points on
Lightning, 6.3 above on Nano, and 6.5 below on Super; on LiveMCPBench 3.5
above on Lightning, level on Nano, and 4.2 below on Super. We read these
two suites as competitive, not as a ranking.

\subsection{What the search contributes}
\label{sec:seed-vs-final}

On Lightning, the hand-written seed program scores 73.1\% on Retail,
68.4\% on Telecom, and 37.9\% on LiveMCPBench (Table~\ref{tab:main}); the
searched program scores 82.7\%, 94.4\%, and 50.5\%, a gain of 9.6, 26.0,
and 12.6 points (59.8\% to 75.9\% averaged over the suites). The search
added rules to all three blocks (Appendix~\ref{app:program}).

\subsection{Ablation: no traces, no primitives}
\label{sec:ablations}

One variant removes both things the traces give the method: the expert
actions that score the search, and the primitives Claude 5 Opus writes from
the clustered traces. In their place Claude Sonnet 4.5 proposes the program
itself. Both versions start from the same hand-written seed, which scores
73.1\% on Retail and 68.4\% on Telecom on Lightning. The proposer keeps a
working memory of its own, a list of findings and rules
that it updates after each proposed program is scored, and at each round
proposes edits to the current program; programs are scored on task outcome
on the training tasks, pass or fail as decided by each task's own
end-state check; nothing is compared with expert actions, and the proposer's
own judgement is never used as a score. The run made 25 proposer calls for
telecom (518,200 prompt and 11,650 completion tokens) and 31 for retail
(539,400 and 15,500).

\begin{center}\small\setlength{\tabcolsep}{8pt}
\begin{tabular}{@{}lrr@{}}
\toprule
\textbf{Lightning} & \textbf{Retail} & \textbf{Telecom} \\
\midrule
\textsc{PLVR}: A$^*$ over mined rules and primitives, reference-action objective & 82.7 & 94.4 \\
Sonnet proposes the program, no traces, no primitives, outcome-scored      & 78.0 & 75.9 \\
\bottomrule
\end{tabular}
\end{center}

Without the traces and the primitives, the proposed program reaches 78.0\%
on Retail and 75.9\% on Telecom on Lightning, against 82.7\% and 94.4\%:
4.7 points lower on Retail and 18.5 lower on Telecom. On Telecom it is
above the seed (68.4\%) and the fine-tune (48.0\%) but well below the
program built from the teacher's actions. This bears on the procedure
thesis in two ways. The seed alone takes Lightning from 47.4\% to 68.4\%
on Telecom, and a frontier model's edits from its own knowledge of the
domain, with no traces, take it to 75.9\%, so a large part of what the
student lacks is procedure that can be written down without the teacher.
The remaining 18.5 points come only when the procedure is taken from the
teacher's actions, so the traces supply procedure that neither the seed
nor the frontier model's prior knowledge does.

\subsection{Tokens and compute}
\label{sec:cost}

\begin{table}[!htbp]
\centering
\caption{Tokens per task, mean over the $\tau^2$-Bench evaluation runs,
counted with each model's own tokenizer. Vanilla, SFT, and GEPA rows pool
the $\tau^2$ runs they were measured on (task counts shown); \textsc{PLVR},
ReAct, and sample-and-judge rows pool Retail and Telecom (228 tasks). For
\textsc{PLVR} the student's tokens and Granite-4.2-8B's tokens (summed over
the primitive calls in a task) are shown separately; for ReAct and
sample-and-judge the Granite columns are the loop or the sampling and
judging, and output includes reasoning tokens; their student columns are
the student's own calls under the loop or across the five samples. The
program's student output count is from Telecom, where it was counted
exactly.}
\label{tab:cost}
\small
\setlength{\tabcolsep}{5pt}
\begin{tabular}{@{}llrrrrrr@{}}
\toprule
\hdrrow & & & \multicolumn{2}{c}{Student} & \multicolumn{3}{c}{Granite-4.2-8B} \\
\hdrrow \cmidrule(lr){4-5} \cmidrule(lr){6-8}
\hdrrow Student & Arm & Tasks & Output & Input & Calls & Output & Input \\
\midrule
Nemotron 3.5 Lightning & Vanilla       & 278 & 2,323 & 45,907 & -- & -- & -- \\
                       & SFT           & 278 & 1,646 & 37,704 & -- & -- & -- \\
                       & GEPA          & 278 & 2,356 & 45,137 & -- & -- & -- \\
                       & ReAct & 228 & 10,700 & 176,800 & 100 & 29,100 & 492,000 \\
                       & Sample-and-judge & 228 & 25,700 & 532,700 & 99 & 29,500 & 634,000 \\
\oursrow               & \textsc{PLVR} & 228 & 1,300 & 249,200 & 112 & 9,000 & 377,000 \\
\midrule
Nemotron 3 Nano        & Vanilla       & 270 & 3,997 & 76,916 & -- & -- & -- \\
                       & SFT           & 278 & 4,518 & 83,366 & -- & -- & -- \\
                       & GEPA          & 181 & 3,546 & 40,498 & -- & -- & -- \\
                       & ReAct & 228 & 15,100 & 183,800 & 107 & 35,700 & 522,000 \\
                       & Sample-and-judge & 228 & 77,300 & 612,200 & 102 & 87,500 & 734,000 \\
\oursrow               & \textsc{PLVR} & 228 & 2,200 & 255,900 & 100 & 8,000 & 326,500 \\
\midrule
Nemotron 3 Super       & Vanilla       & 278 & 2,672 & 63,948 & -- & -- & -- \\
                       & SFT           & 278 & 2,640 & 56,071 & -- & -- & -- \\
                       & GEPA          & 278 & 2,747 & 63,682 & -- & -- & -- \\
                       & ReAct & 228 & 9,400 & 175,800 & 97 & 26,400 & 479,000 \\
                       & Sample-and-judge & 228 & 26,400 & 574,800 & 105 & 30,400 & 686,000 \\
\oursrow               & \textsc{PLVR} & 228 & 1,400 & 248,100 & 117 & 9,700 & 407,500 \\
\bottomrule
\end{tabular}
\end{table}

Table~\ref{tab:cost} gives tokens per task on the two $\tau^2$ domains.
Under the program the student itself reads 248,100--255,900 input tokens per
task and writes 1,300--2,200 output tokens; Granite-4.2-8B adds 100--117
primitive calls per task, reading 326,500--407,500 tokens and writing
8,000--9,700 in total. The untreated student reads 45,907--76,916 input
tokens per task and writes 2,323--3,997; fine-tuning and GEPA are within
the same range. The two token-matched arms make 97--107 Granite calls per
task, against 100--117 for the program, and read 479,000--734,000 Granite
input tokens against the program's 326,500--407,500.

\paragraph{FLOPs.} We count $2 \times$ active parameters $\times$ tokens for
inference and $6 \times$ active parameters $\times$ token-passes for
training, taking 3B active parameters for Lightning and Nano, 12B for Super,
and 8B for Granite-4.2-8B. The input counts in Table~\ref{tab:cost} are
cumulative over calls; Granite's calls run with prefix caching and about
80\% of its input tokens are served from cache, which we cost at zero. The
student's cached fraction was not measured, so student input is costed in
full for every arm. Per task, a vanilla run costs $2.9 \times 10^{14}$,
$4.9 \times 10^{14}$, and $1.6 \times 10^{15}$ FLOPs on Lightning, Nano,
and Super; a run under the program costs $2.9 \times 10^{15}$,
$2.7 \times 10^{15}$, and $7.5 \times 10^{15}$, of which Granite is
$1.4 \times 10^{15}$, $1.2 \times 10^{15}$, and $1.5 \times 10^{15}$; the
ratios are 9.8, 5.6, and 4.7.
Table~\ref{tab:build} gives the cost of producing each arm. The fine-tuning
sweep is 24 runs per student (Appendix~\ref{app:distill}), $8.9 \times
10^{8}$ token-passes each; the search is 383,783 Granite calls and 1.87 billion tokens over the three
blocks (257,557 calls and 1,159M tokens for telecom, 96,814 and 605M for
retail, 29,412 and 104M for tool-search; Appendix~\ref{app:hyperparams}),
run once for all three students. Mining the rule library took Claude
Sonnet 4.5 16 calls: 6 for the tool-search traces (47,047 prompt and
17,685 completion tokens) and 10 for the telecom traces (239,600 and
29,500). It is an API model whose size is not public, so its FLOPs are not
counted in Table~\ref{tab:build}.

\begin{table}[!htbp]
\centering
\caption{Cost of producing each arm, in FLOPs. Fine-tuning is per student;
the search is run once and its program is used by every student. The
selected-run figure is the one fine-tune of Table~\ref{tab:main}; the sweep
is all 24 runs that selection drew on.}
\label{tab:build}
\small
\begin{tabular}{@{}lrrr@{}}
\toprule
\hdrrow & Lightning (3B) & Nano (3B) & Super (12B) \\
\midrule
SFT, selected run only      & $7.7 \times 10^{17}$ & $7.7 \times 10^{17}$ & $3.1 \times 10^{18}$ \\
SFT, full sweep (24 runs)   & $1.6 \times 10^{19}$ & $1.6 \times 10^{19}$ & $6.4 \times 10^{19}$ \\
\oursrow \textsc{PLVR} search, Granite scoring, all three blocks & \multicolumn{3}{c}{$3.0 \times 10^{19}$ (once)} \\
\oursrow \textsc{PLVR} rule mining (Claude Sonnet 4.5) & \multicolumn{3}{c}{16 calls, 287k prompt and 47k completion tokens (once; FLOPs not counted)} \\
\bottomrule
\end{tabular}
\end{table}

\FloatBarrier
\section{Discussion}
\label{sec:discussion}

\paragraph{How much is procedure depends on the domain.} The three suites
give different answers. On Telecom the program with Granite-4.2-8B as the
student scores 91.2\%, every Nemotron student under it 93.3--94.4\%, and no
other arm above 58.2\%; the procedure is nearly the whole result, and on
this suite an 8B model is enough to run it. The result is in-domain
compositional generalisation: the program is built from synthetic tasks in
the benchmark's own environment, with its tools and policies, and is run on
the benchmark's test tasks, which it has not seen; it is not a claim about
transfer to another domain. Telecom is also where the program's student runs
longest, about 40 turns per task against about 9 on Retail. On Retail the
program with Granite as the student scores 82.5\%, within 2.7 points of the
Nemotron students, so there too the choice of student matters little under
the program, though the program's margin over the baselines is small. On
LiveMCPBench the program with Granite as the student reaches 42.1\% and the
Nemotron students are 3.2 to 12.6 points ahead of it; the student's tool
choice and argument binding still count. We read the part of the gap
between base models that the program closes as an estimate of how much of
that gap is procedure; the program also carries the frontier models' prior
knowledge (\S\ref{sec:ablations}), so the estimate is an upper bound.

\paragraph{The same program for every student.} The program is the same
object in every row: the primitives run on Granite-4.2-8B, the search was
run once, and no part of it is tuned to a student. Under it, the three
Nemotron students land within 1.1 points of each other on Telecom and 3.5
on Retail, and Granite as the student lands within 3.2 points of them on
both; on LiveMCPBench they stay 9.4 points apart. Where the base model still
matters is where the student is doing part of the work.

\paragraph{What it costs.} The cost of the program is in inference tokens:
the student reads 3.3 to 5.4 times as many input tokens as under the other
arms, and Granite's calls add more than that again; per task this is 4.7 to
9.8 times the FLOPs of a vanilla run. Producing the program cost
$3.0 \times 10^{19}$ FLOPs in Granite scoring, counting every Granite
token in full and excluding the rule-mining calls, once for all three
students; producing one fine-tune
cost $7.7 \times 10^{17}$ to $3.1 \times 10^{18}$ for the selected run and
$1.6 \times 10^{19}$ to $6.4 \times 10^{19}$ for the sweep that selected it,
per student. No weights change, and the search was run once and the
resulting program is shared by every student.

\paragraph{Limitations.} The gain is not attributed to the traces alone.
The hand-written seed, Claude 5 Opus (which writes the primitives), and
Claude Sonnet 4.5 (which mines the rules) all contribute, and neither
frontier model is a blank slate about customer service or tool use; the
ablation that removes the traces and the primitives still reaches 75.9\% on
Telecom (\S\ref{sec:ablations}), and that ablation covers Retail and
Telecom only. All cells but the program's Telecom cells are single runs, so
differences of a few points are not resolved. The synthetic training tasks
share the $\tau^2$ Retail and Telecom domains with the test suites and the
LiveMCPBench format with LiveMCPBench, so domain knowledge can pass from
the training set into the primitives, the rules, and the fine-tunes. The
training tasks and their answers are different from the test tasks; the
authors verified this.

\section{Conclusion}
\label{sec:conclusion}

We compiled a teacher's procedure into a program rather than into a
student's weights: primitives authored by a frontier model from clusters of
the teacher's traces, each run by a small frozen model and verified as it
runs, assembled by A$^*$ over a rule library mined from the same traces
under a reference-action objective, and used at inference to write the
student's chain of thought or to check the actions the student proposes.
Run in front of three Nemotron base models and on its own, the one program
estimates how much of each domain is procedure. On $\tau^2$ Telecom the
estimate is that nearly all of it is: Granite-4.2-8B as the student goes
from 43.0\% to 91.2\%, every Nemotron student reaches 93.3--94.4\%, and
no other arm passes 58.2\%; on that suite an 8B model is enough. This is
in-domain compositional generalisation, from synthetic tasks in the
benchmark's environment to its unseen test tasks. On $\tau^2$
Retail the 8B student under the program is within 2.7 points of the
Nemotron students; on LiveMCPBench the Nemotron students are 3 to 13 points
ahead of it. On those two suites the program is competitive with the best
baseline. No weights are updated, the program is the same for
every base model, and the spread between base models shrinks where the
program does the work. The program was produced once, at
$3.0 \times 10^{19}$ FLOPs, and costs 5 to 10 times a vanilla run to
execute. Its output is a program rather than a checkpoint: it can be read
at every step, it is the same for every base model, and it can be revised
one primitive at a time. The next test is cross-domain transfer: a program
built in one environment run in another.

\section*{Acknowledgements}
We thank Ram Shanmugam and Jay Vaghasiya for their help with the experiments
and the paper.

\section*{Reproducibility Statement}
The method is specified in \S\ref{sec:method}. The synthetic training set, the
kept teacher traces, the primitive prompts, the programs, the search settings,
and the baseline construction are in the appendices. The synthetic
dataset, the program, the primitives, and the search code will be released
following peer review.

\bibliography{iclr2027_conference}
\bibliographystyle{iclr2027_conference}

\appendix

\section{Search settings}
\label{app:hyperparams}

\begin{center}\footnotesize\setlength{\tabcolsep}{5pt}
\begin{tabular}{@{}lp{0.70\linewidth}@{}}
\toprule
\textbf{Item} & \textbf{Value} \\
\midrule
state & the seed plus a set of rules \\
rule library & 125 rules mined once by Claude Sonnet 4.5: 61 from the tool-search traces, 64 from the telecom traces (Appendix~\ref{app:mined}); the same library for every block, retail included \\
successor & insert one rule, or one group of rules, from the library into the current program; kept only if the program still compiles and runs without errors \\
final pick & best task outcome on the training tasks among the programs reached \\
Granite scoring cost & telecom 257,557 calls, 1,159M tokens; retail 96,814 calls, 605M tokens; tool-search 29,412 calls, 104M tokens \\
mining cost & tool-search 6 calls, 47,047 prompt and 17,685 completion tokens; telecom 10 calls, 239,600 and 29,500 \\
objective & $f = |P| + 100\,d$; $|P|$ = rules in the program, $d$ = mean distance to the expert actions over the search set (\S\ref{sec:process}) \\
\bottomrule
\end{tabular}
\end{center}

\paragraph{Programs evaluated.}

\begin{center}\footnotesize\setlength{\tabcolsep}{6pt}
\begin{tabular}{@{}ll@{}}
\toprule
\textbf{Block} & \textbf{Programs scored} \\
\midrule
Telecom     & 32 \\
Retail      & 87 \\
Tool-search & 9: seed, six group edits, their merge, final \\
\bottomrule
\end{tabular}
\end{center}

\section{Seed and final programs}
\label{app:program}

The program is three blocks, one per subset of the training set: a telecom
block for troubleshooting, a retail block for multi-turn tool matching, and
a tool-search block for the LiveMCPBench format. Each block is given here
twice: the hand-written seed, and the final program after the search. In the
final listings, a line with a leading \texttt{+} is a rule the search
added from the mined library, written as the library gives it, with the
library id for tool-search rules (Appendix~\ref{app:mined}); unmarked
lines are from the seed. The retail block's added rules come from the same
library, mined from the telecom and tool-search traces.

\subsection{Telecom block (troubleshooting)}

\paragraph{Seed.}

\begin{lstlisting}[style=prompt]
split the message into requests
for each request that reports a problem:
    walk the policy's checklist in its written order
    for each fix rule:
        read the current state of the system
        do not repeat a fix that is already in place
    re-run the policy's check; stop when it passes
drop duplicate actions
\end{lstlisting}

\paragraph{Final.}

\begin{lstlisting}[style=prompt]
split the message into requests
for each request that reports a problem:
+   identify first: looks up the user before anything else
+   find the right record: reads all records to match the user's identifier
    go through the rules' diagnostic steps in their written order
+   one step at a time: executes a procedure written as a list of steps one
+       step at a time
    for each rule of the form "if <condition>, apply <fix>":
+       check the problem early: confirms the reported problem before
+           attempting fixes
+       customer acts on their device: never calls user-side tools directly
        read the current state
+       read before changing a record: never changes settings without reading
+           them first
        skip a fix whose effect is already in place
+       argument names the target: an identifying argument names the target,
+           not the change to make
+       refusal cites the rule: a refusal cites a value the rule's text
+           mentions
+   before a state-changing action:
+       quantities from records: every quantity is user-stated or computed
+           from records
+       target in the named state: when the request names a state, the target
+           chosen is in that state
+       verify payment: confirms the bill shows as paid before proceeding
+   after a state-changing action, request the follow-up step at once:
+       restart after a settings reset, restart after resuming,
+       toggle on the device after the account
+   escalate after a failed attempt: escalates only after the action was
+       attempted and refused or failed
+   dates computed in code: computes time conditions in code against the date
+       the rules give
+   ended contract to a human: transfers without trying to resume service
    re-run the rules' success check; stop when it passes
remove duplicate actions
\end{lstlisting}

\subsection{Retail block (multi-turn tool matching)}

The base model
proposes each turn's action and the block checks it before it goes out.

\paragraph{Seed.}

\begin{lstlisting}[style=prompt]
each turn:
    proposal = base model's next action (a lookup, a message or a change)

    unknown tool or malformed id            -> rejected, proposed again
    a change: every value must be in a record that was read or in the
        customer's words
    a change: customer identified, owns the record, policy rule holds;
        else refuse
    a change goes out only after the customer's yes to its values
    a change already made is not made again
    "done" claimed without the call         -> retry
    no offers the customer did not ask for

    if nothing above fires -> the proposal goes out unchanged
\end{lstlisting}

\paragraph{Final.}

\begin{lstlisting}[style=prompt]
each turn:
    proposal = base model's next action (read, message, or state change)

    # validity
    the action must exist and its arguments must be well-formed
+   fix after validation error: retries with parameters corrected from the
+       error (M42)

    # grounding: every value has a source
    every value in a state change comes from an observed record or the user
+   fill from the record: takes a value the user did not give from the record
+       and confirms it
+   computed values are own work: computed figures come from its own
+       calculation on retrieved data (M49)
+   carry identifiers forward: uses an identifier from one result in the next
+       call (M15)

    # reference: resolve what the user means before acting
+   read all before comparing: reads every source before a comparison or join
+       (M47)
+   preserve existing content: reads what exists, then writes old and new
+       content together (M17)
+   superlatives by comparison: retrieves all values, then compares to find
+       the top one (M13)
+   user's fallback: applies the user's stated fallback when the request
+       cannot be met

    # permission: the rules decide, not the base model
    a state change requires: user identified, owns the target, rules allow it
+   no tools for simple logic: never searches for a tool to do arithmetic,
+       parsing or filtering (M05)
+   permitted equivalent: uses an equivalent action a rule permits when the
+       rule blocks the requested one
+   refuse only on a rule: refuses only when the rule check produced the
+       refusal

    # consent
    a state change goes out only after the user agrees to its exact values
+   explicit agreement: agreement is explicit, covers one state change, and
+       is not inferred from a choice among options or a qualified yes
+   confirm once: does not confirm an agreed state change a second time

    # ordering
+   agreed change first: carries out an agreed state change before anything
+       else
+   irreversible last: does irreversible actions last, after telling the user
+       they are final
+   correction everywhere: applies a correction everywhere the corrected
+       value appears
+   drop a no-op: drops an action that would change nothing

    # restraint
    a state change already made is not repeated
    a claim of completion without the action is retried
    nothing is offered that the user did not ask for
+   checks stay internal: does not surface internal checks to the user as
+       questions
+   stop after answering: takes no further action after the final answer
+       (M03)
+   no early escalation: does not escalate while the user is still willing to
+       continue

    if nothing above fires -> the proposal goes out unchanged
\end{lstlisting}

\subsection{Tool-search block (LiveMCPBench format)}

\paragraph{Seed.}

\begin{lstlisting}[style=prompt]
split the task into requests
for each request:
    retrieve candidate tools for the request
    pick one, fill its parameters
    call it
    check whether the request is achieved
answer
\end{lstlisting}

\paragraph{Final.}

\begin{lstlisting}[style=prompt]
split the task into requests
+ one request if the split fails: treats the whole task as one request when
+     the split gives nothing usable (M56)
+ complete answer: the final answer includes everything the task asked for
+     (M02)
for each request:
    retrieve candidate tools for the request
+   search in the tool's form: uses the form the search tool requires and
+       never repeats a search unchanged (M57)
    pick one, fill its parameters
+   fix after validation error: retries with parameters corrected from the
+       error (M42)
+   exact text: uses text the task specifies word for word (M20)
+   read before use: reads a source before doing anything with its contents
+       (M01)
+   read before editing: reads a source before making a precise edit (M45)
+   call values from the task or results: values in a call come from the task
+       or a tool result (M58)
    call it
+   empty is not success: treats an empty result, or one that fails for every
+       item, as not a success (M59)
+   alternative after error: tries a different approach, not the same call
+       again (M14)
+   different tool after empty result: moves to a more direct query or a
+       known identifier (M44)
+   drop a dead source: does not use again a source that fails to connect or
+       lacks the tool (M60)
+   specialised lookup: uses the dedicated lookup tool for an identifier
+       system (M40)
    check whether the request is achieved
+   all calls then one answer: makes every needed call, then gives a combined
+       answer (M55)
+   report the output: the answer names where the output went and summarises
+       it (M07)
+   gather before writing: finishes all data gathering before writing an
+       output (M06)
+   stop after answering: takes no further action after the final answer
+       (M03)
+   budget in calls: counts the planner budget in calls per action, not
+       seconds (M61)
answer
+ computed values are own work: computed figures come from its own
+     calculation on retrieved data (M49)
\end{lstlisting}

\paragraph{Seed against final.}

\begin{center}\footnotesize\setlength{\tabcolsep}{8pt}
\begin{tabular}{@{}lrr@{}}
\toprule
\textbf{Nemotron 3.5 Lightning} & \textbf{Seed} & \textbf{Final} \\
\midrule
$\tau^2$ Retail   & 73.1\% & 82.7\% \\
$\tau^2$ Telecom  & 68.4\% & 94.4\% \\
LiveMCPBench       & 37.9\% & 50.5\% \\
\bottomrule
\end{tabular}
\end{center}

\section{Fine-tuning baselines}
\label{app:distill}

The fine-tuning arm of the comparison is supervised fine-tuning of each Nemotron
student on exactly the kept teacher traces of \S\ref{sec:data}. This appendix
gives its construction, sweep, and selection.

\subsection{What is matched}

\paragraph{Data.} Matched exactly: both arms see the 539 kept teacher traces
of \S\ref{sec:data}, and nothing else. No arm sees $\tau^2$-Bench's official
test sets or LiveMCPBench.

\paragraph{Rows.} Fine-tuning uses the kept teacher traces of the training
split of each subset (Table~\ref{tab:trainset}): retail, one row per task;
telecom and tool-search, one row per assistant turn, since those tasks are
multi-turn. Validation uses the traces of the held-aside split. Maximum
length is 16,384 tokens and no row was truncated in any run.

\paragraph{Models.} The three Nemotron students, fine-tuned directly:
Nemotron 3.5 Lightning, Nemotron 3 Nano, Nemotron 3 Super.

\subsection{Models}

\begin{center}\footnotesize\setlength{\tabcolsep}{5pt}
\begin{tabular}{@{}ll@{}}
\toprule
\textbf{Student} & \textbf{Checkpoint} \\
\midrule
Nemotron 3 Nano & \texttt{nvidia/NVIDIA-Nemotron-3-Nano-30B-A3B-BF16} \\
Nemotron 3.5 Lightning & \texttt{nvidia/NVIDIA-Nemotron-3.5-Lightning-30B-A3B-BF16} \\
Nemotron 3 Super & \texttt{nvidia/NVIDIA-Nemotron-3-Super-120B-A12B-BF16} \\
\bottomrule
\end{tabular}
\end{center}

\subsection{The variants}

Every row shares a chat layout: the system instruction is identical throughout
and names no primitive; the user message carries the environment state, tool
schemas, and numbered turns exactly as the teacher saw them. The arms differ
only in the assistant target. The \emph{outcome} arm's target is the final
reply with the tool calls only. The \emph{process} arm's target is the
teacher's reasoning as a thinking block followed by the final calls. A
third cell, \emph{process+outcome}, has the same target as the process arm
and was swept separately; the sweep for all three cells is reported below.

\subsection{Training and sweep}

\begin{center}\footnotesize\setlength{\tabcolsep}{5pt}
\begin{tabular}{@{}ll@{}}
\toprule
\textbf{Setting} & \textbf{Value} \\
\midrule
method & LoRA \\
loss & cross-entropy on assistant tokens only \\
optimiser & Adam \\
schedule & linear decay to zero, $\mathrm{lr} \times (1 - \mathrm{step}/\mathrm{total\_steps})$ \\
batch & 8 \\
max length & 16{,}384 tokens \\
seed & 0 \\
validation sampling & temperature 0; 8{,}192 max tokens \\
swept & lr $\in \{1, 3, 10, 30\} \times 10^{-5}$; rank $\in \{8, 32\}$; 6 epochs (a few 1 or 8), checkpointed every epoch, epoch selected post hoc \\
runs & 72, 24 per student (8 configurations per cell) \\
\bottomrule
\end{tabular}
\end{center}

\subsection{Selection}

Selection maximises \texttt{val\_solved} on the held-aside traces of the
three subsets, breaking ties by lower validation NLL. An instance is solved
when the predicted calls block exactly matches the ground truth (no type
errors, loss 0). Selections:

\begin{center}\footnotesize\setlength{\tabcolsep}{4pt}
\begin{tabular}{@{}llrrr@{}}
\toprule
\textbf{Cell} & \textbf{Student} & \textbf{lr} & \textbf{rank} & \textbf{epoch} \\
\midrule
process          & Nano      & $3\times10^{-4}$ & 32 & 6 \\
outcome          & Nano      & $3\times10^{-4}$ & 8  & 1 \\
process+outcome  & Nano      & $1\times10^{-4}$ & 32 & 2 \\
process          & Lightning & $1\times10^{-4}$ & 8  & 2 \\
outcome          & Lightning & $1\times10^{-4}$ & 32 & 6 \\
process+outcome  & Lightning & $1\times10^{-4}$ & 8  & 3 \\
process          & Super     & $3\times10^{-4}$ & 8  & 3 \\
outcome          & Super     & $1\times10^{-4}$ & 32 & 1 \\
process+outcome  & Super     & $1\times10^{-5}$ & 32 & 1 \\
\bottomrule
\end{tabular}
\end{center}

\section{Extended methodology and analysis}

\subsection{Primitive extraction: full procedure}
\label{app:extraction}

\subsubsection{Segmentation and clustering}

The kept traces are split into 3,141 segments, each typed as a reason, a
call, a message, or a question. Segmentation walks each trace's message list. Only assistant messages
produce segments, and each produces one segment per thing it emitted: a
non-empty reasoning field gives one \emph{reason} segment; each entry in the
tool-call list gives one \emph{call} segment; non-empty content gives one
\emph{message} segment, or a \emph{question} segment if it ends in a
question mark. User and tool messages produce no segments; they accumulate
into the context (the last user message and the pending tool results) that
is attached to the next assistant segment, and the pending results are
cleared after each agent turn. The cut is therefore per emitted product
rather than per turn: one assistant turn with reasoning, two calls, and a
closing message becomes four segments. A segment is
represented by the agent's output at that point only, truncated to 2,000
characters: for a reason, the reasoning span; for a call, the literal
\texttt{name(\{json args\})}; for a message or question, the message text.
The surrounding context (user message, tool results, turn, domain, function)
rides on the segment and reaches the frontier-model stages but is never
embedded. Segments are embedded with \texttt{text-embedding-3-small}
(1,536 dimensions, L2-normalised); all similarities are cosine. Segments are clustered with
$k$-means within each kind, giving 35 clusters: 15 reason (876 segments), 8
call (1,064), 9 message (1,159), and 3 question (42).

\subsubsection{From clusters to primitives}

Clusters locate which segments a primitive lives in; they do not separate
which decision that segment witnesses. Claude 5 Opus post-processes the
clusters: it decides which operations they show and writes each primitive's
prompt, input--output signature, verifier, and fallback. The 37 primitives
are listed below by group, with the subsets they are used in (T = telecom,
M = tool-search, R = retail), the number of segments assigned to them, and
the clusters those segments come from.

\paragraph{Understanding the message.}
\begin{center}\footnotesize\setlength{\tabcolsep}{4pt}
\begin{tabular}{@{}lllp{0.58\linewidth}@{}}
\toprule
\textbf{Primitive} & \textbf{Used in} & \textbf{Seg.} & \textbf{Clusters (segments each)} \\
\midrule
Split into requests & T, M & 12 & reason9 (7), message0 (3), call5 (2) \\
Classify one request & T, M & 9 & reason8 (4), reason2 (2), message0 (2), reason12 (1) \\
Order requests & T, M & 30 & reason17 (10), reason9 (8), reason11 (5), message0 (4), reason15 (3) \\
Route a reply & T & 1 & call13 (1) \\
Device reply & T & 195 & message8 (68), reason16 (26), message2 (25), message4 (24), reason2 (21), call2 (8), question2 (5), question5 (5), call11 (4), call13 (3), message9 (2), reason9 (2), reason14 (1), call9 (1) \\
Check a claim & T & 2 & reason10 (2) \\
Parts & M & 25 & call7 (23), call5 (2) \\
\bottomrule
\end{tabular}
\end{center}

\paragraph{Finding the tools.}
\begin{center}\footnotesize\setlength{\tabcolsep}{4pt}
\begin{tabular}{@{}lllp{0.58\linewidth}@{}}
\toprule
\textbf{Primitive} & \textbf{Used in} & \textbf{Seg.} & \textbf{Clusters (segments each)} \\
\midrule
Servers for the task & M & 59 & call1 (26), call5 (17), call10 (16) \\
Search phrases & M & 7 & call1 (5), call5 (2) \\
Widen tools & M & 4 & reason5 (2), call5 (2) \\
Pick the function & T, M & 35 & call5 (17), call1 (16), call10 (2) \\
Plan steps & T, M & 26 & reason17 (13), reason5 (6), reason9 (4), message0 (1), reason11 (1), reason8 (1) \\
\bottomrule
\end{tabular}
\end{center}

\paragraph{Building the call.}
\begin{center}\footnotesize\setlength{\tabcolsep}{4pt}
\begin{tabular}{@{}lllp{0.58\linewidth}@{}}
\toprule
\textbf{Primitive} & \textbf{Used in} & \textbf{Seg.} & \textbf{Clusters (segments each)} \\
\midrule
Fill the parameters & T, M, R & 30 & call9 (11), call5 (8), call10 (6), message1 (2), call1 (2), reason11 (1) \\
Where a value comes from & T, M & 35 & message0 (15), call9 (6), reason9 (3), reason3 (3), reason15 (1), message7 (1), reason17 (1), reason13 (1), message4 (1), reason8 (1), reason2 (1), reason1 (1) \\
How to get a missing parameter & T, M & 27 & reason2 (26), reason1 (1) \\
Question for a missing parameter & T, R & 20 & reason2 (13), reason5 (3), reason11 (2), reason3 (2) \\
Device action arguments & T & 2 & call5 (2) \\
Requests covered by the same call & T & 6 & reason9 (2), reason17 (1), message0 (1), reason15 (1), reason11 (1) \\
Price terms & T & 9 & message5 (4), reason5 (1), message9 (1), question4 (1), reason9 (1), message7 (1) \\
Which arguments failed & T, M & 1 & call10 (1) \\
\bottomrule
\end{tabular}
\end{center}

\paragraph{Checking the policy.}
\begin{center}\footnotesize\setlength{\tabcolsep}{4pt}
\begin{tabular}{@{}lllp{0.58\linewidth}@{}}
\toprule
\textbf{Primitive} & \textbf{Used in} & \textbf{Seg.} & \textbf{Clusters (segments each)} \\
\midrule
Functions a policy section governs & T, R & 14 & reason5 (9), reason4 (4), reason11 (1) \\
Is the rule a condition & T, R & 3 & reason14 (1), reason16 (1), reason9 (1) \\
Is the rule about this action & T & 17 & reason17 (8), reason5 (4), reason9 (2), message0 (2), reason11 (1) \\
Is the condition met & T, R & 7 & reason9 (3), reason17 (2), reason3 (1), reason8 (1) \\
User's own premise or condition & T & 36 & reason17 (16), reason11 (8), reason9 (5), reason5 (5), reason8 (2) \\
Alternatives when blocked & T & 10 & reason5 (9), reason9 (1) \\
Reply to a confirmation & T & 5 & message0 (4), reason17 (1) \\
\bottomrule
\end{tabular}
\end{center}

\paragraph{Troubleshooting conditions.}
\begin{center}\footnotesize\setlength{\tabcolsep}{4pt}
\begin{tabular}{@{}lllp{0.58\linewidth}@{}}
\toprule
\textbf{Primitive} & \textbf{Used in} & \textbf{Seg.} & \textbf{Clusters (segments each)} \\
\midrule
Copy the rule's condition & T & 1 & reason16 (1) \\
Does one reading show the condition & T & 1 & reason14 (1) \\
Does the condition hold now & T & 11 & reason17 (3), reason9 (3), message0 (2), reason14 (1), reason15 (1), reason4 (1) \\
Is the condition true by the quote alone & T & 5 & reason17 (1), reason3 (1), reason5 (1), message4 (1), message0 (1) \\
Is the setting already as wanted & T & 14 & message2 (5), reason4 (4), message8 (3), reason5 (2) \\
Is the problem fixed & T & 29 & reason4 (9), message2 (7), message9 (4), message3 (4), message1 (2), question2 (2), reason17 (1) \\
\bottomrule
\end{tabular}
\end{center}

\paragraph{Finishing.}
\begin{center}\footnotesize\setlength{\tabcolsep}{4pt}
\begin{tabular}{@{}lllp{0.58\linewidth}@{}}
\toprule
\textbf{Primitive} & \textbf{Used in} & \textbf{Seg.} & \textbf{Clusters (segments each)} \\
\midrule
Is the request achieved & M & 3 & reason17 (1), reason4 (1), message7 (1) \\
Audit & M & 171 & message1 (118), message5 (31), message3 (20), message0 (2) \\
Can the question be answered & T, M & 2 & reason5 (1), reason9 (1) \\
Write the reply & T, M & 9 & message0 (7), call5 (2) \\
\bottomrule
\end{tabular}
\end{center}

\subsubsection{Clusters, and the primitives each carries}

\begin{center}\footnotesize\setlength{\tabcolsep}{4pt}
\begin{longtable}{@{}llrlrp{0.52\linewidth}@{}}
\toprule
\textbf{Cluster} & \textbf{Kind} & \textbf{Size} & \textbf{Domains} & \textbf{\#} & \textbf{Primitives} \\
\midrule
\endhead
\bottomrule
\endfoot
reason9 & reason & 24 & telecom & 14 & Alternatives when blocked, Can the question be answered, Device reply, Does the condition hold now, Is the condition met, Is the rule a condition, Is the rule about this action, Order requests, Plan steps, Price terms, Requests covered by the same call, Split into requests, User's own premise or condition, Where a value comes from \\
message0 & message & 236 & tool-search, retail, telecom & 12 & Audit, Classify one request, Does the condition hold now, Is the condition true by the quote alone, Is the rule about this action, Order requests, Plan steps, Reply to a confirmation, Requests covered by the same call, Split into requests, Where a value comes from, Write the reply \\
reason17 & reason & 41 & retail & 12 & Does the condition hold now, Is the condition met, Is the condition true by the quote alone, Is the problem fixed, Is the request achieved, Is the rule about this action, Order requests, Plan steps, Reply to a confirmation, Requests covered by the same call, User's own premise or condition, Where a value comes from \\
reason5 & reason & 117 & tool-search, telecom & 11 & Alternatives when blocked, Can the question be answered, Functions a policy section governs, Is the condition true by the quote alone, Is the rule about this action, Is the setting already as wanted, Plan steps, Price terms, Question for a missing parameter, User's own premise or condition, Widen tools \\
call5 & call & 128 & tool-search & 9 & Device action arguments, Fill the parameters, Parts, Pick the function, Search phrases, Servers for the task, Split into requests, Widen tools, Write the reply \\
reason11 & reason & 102 & telecom & 8 & Fill the parameters, Functions a policy section governs, Is the rule about this action, Order requests, Plan steps, Question for a missing parameter, Requests covered by the same call, User's own premise or condition \\
reason2 & reason & 89 & telecom & 5 & Classify one request, Device reply, How to get a missing parameter, Question for a missing parameter, Where a value comes from \\
reason4 & reason & 88 & tool-search, telecom & 5 & Does the condition hold now, Functions a policy section governs, Is the problem fixed, Is the request achieved, Is the setting already as wanted \\
reason8 & reason & 70 & tool-search, retail & 5 & Classify one request, Is the condition met, Plan steps, User's own premise or condition, Where a value comes from \\
call1 & call & 204 & tool-search & 4 & Fill the parameters, Pick the function, Search phrases, Servers for the task \\
call10 & call & 179 & tool-search & 4 & Fill the parameters, Pick the function, Servers for the task, Which arguments failed \\
reason14 & reason & 23 & tool-search, telecom & 4 & Device reply, Does one reading show the condition, Does the condition hold now, Is the rule a condition \\
reason15 & reason & 38 & retail & 4 & Does the condition hold now, Order requests, Requests covered by the same call, Where a value comes from \\
reason3 & reason & 36 & retail & 4 & Is the condition met, Is the condition true by the quote alone, Question for a missing parameter, Where a value comes from \\
call9 & call & 51 & telecom & 3 & Device reply, Fill the parameters, Where a value comes from \\
message1 & message & 125 & tool-search & 3 & Audit, Fill the parameters, Is the problem fixed \\
message2 & message & 175 & telecom & 3 & Device reply, Is the problem fixed, Is the setting already as wanted \\
message4 & message & 81 & telecom & 3 & Device reply, Is the condition true by the quote alone, Where a value comes from \\
message7 & message & 95 & retail & 3 & Is the request achieved, Price terms, Where a value comes from \\
message9 & message & 197 & tool-search, telecom & 3 & Device reply, Is the problem fixed, Price terms \\
reason16 & reason & 99 & telecom & 3 & Copy the rule's condition, Device reply, Is the rule a condition \\
call13 & call & 38 & telecom & 2 & Device reply, Route a reply \\
message3 & message & 25 & tool-search & 2 & Audit, Is the problem fixed \\
message5 & message & 34 & tool-search & 2 & Audit, Price terms \\
message8 & message & 191 & telecom & 2 & Device reply, Is the setting already as wanted \\
question2 & question & 13 & telecom & 2 & Device reply, Is the problem fixed \\
reason1 & reason & 35 & tool-search, telecom & 2 & How to get a missing parameter, Where a value comes from \\
call11 & call & 38 & telecom & 1 & Device reply \\
call2 & call & 272 & telecom & 1 & Device reply \\
call7 & call & 154 & retail & 1 & Parts \\
question4 & question & 4 & telecom & 1 & Price terms \\
question5 & question & 25 & telecom & 1 & Device reply \\
reason10 & reason & 27 & tool-search & 1 & Check a claim \\
reason12 & reason & 76 & retail & 1 & Classify one request \\
reason13 & reason & 11 & tool-search, retail & 1 & Where a value comes from \\
\end{longtable}
\end{center}

\section{Mined rule libraries}
\label{app:mined}

For the search of \S\ref{sec:search}, Claude Sonnet 4.5 mines a rule
library from the kept traces in a few calls, and the two libraries below
are used together, as one candidate set of 125 rules, for every block,
including the retail block, which has no library of its own: several
extraction calls over batches of traces, then one merge call. Each rule is
a one-line behaviour the expert shows. The tool-search library (61 rules)
took 6 calls (5 extraction, 1 merge; 47,047 tokens in, 17,685 out). The
telecom library (64 rules) took 10 calls (9 extraction, 1 merge; 239,600
tokens in, 29,500 out).

\subsection{Tool-search (LiveMCPBench format), 61 rules}

\begin{center}\footnotesize\setlength{\tabcolsep}{4pt}
\begin{longtable}{@{}llp{0.6\linewidth}@{}}
\toprule
\textbf{Id} & \textbf{Rule} & \textbf{What the expert does} \\
\midrule
\endhead
\bottomrule
\endfoot
M01 & Read before use & Reads a source before doing anything with its contents \\
M02 & Complete answer & The final answer includes everything the task asked for \\
M03 & Stop after answering & Takes no further action after the final answer \\
M04 & Inputs before calculating & Retrieves all input values, then computes locally \\
M05 & No tools for simple logic & Never searches for a tool to do arithmetic, parsing or filtering \\
M06 & Gather before writing & Finishes all data gathering before writing an output \\
M07 & Report the output & The answer names where the output went and summarises it \\
M08 & Only needed fields & Takes only the fields needed from a structured result \\
M09 & All sources first & Retrieves from every source before composing the answer \\
M10 & Direct lookup & Retrieves a named entity directly, without searching first \\
M11 & One fetch per entity & Answers several questions about an entity from one result \\
M12 & One call per entity & Uses a separate call for each entity of the same type \\
M13 & Superlatives by comparison & Retrieves all values, then compares to find the top one \\
M14 & Alternative after error & Tries a different approach, not the same call again \\
M15 & Carry identifiers forward & Uses an identifier from one result in the next call \\
M16 & Build in order & Creates the container, then its structure, then its content \\
M17 & Preserve existing content & Reads what exists, then writes old and new content together \\
M18 & Filter locally & Reads the whole dataset and filters it itself \\
M19 & Count locally & Reads the data and counts it itself \\
M20 & Exact text & Uses text the task specifies word for word \\
M21 & Exact output location & Uses the output location exactly as given \\
M22 & Aggregate after reading & Reads all data, computes, then writes results \\
M23 & Dependent values in order & Gets each input a calculation depends on, in order, then calculates \\
M24 & Inputs with the result & States the inputs, the result and the factor used \\
M25 & Resolve both ends before relating & Resolves both entities before asking for the relation between them \\
M26 & Entity before attribute & Resolves an entity before querying an attribute of it \\
M27 & Context before a dependent value & Gets the context a value depends on, then asks for the value \\
M28 & Read each part & Reads each part of a source separately, then combines \\
M29 & Locate before updating & Finds the items to change, then writes each one \\
M30 & Report the count changed & States how many items were updated \\
M31 & Include the artefact reference & Puts the produced artefact's reference in the answer \\
M32 & Shape data for the consumer & Transforms earlier results into the format the next tool needs \\
M33 & Search then details & Searches, then fetches details by the result's identifier \\
M34 & Check state before changes & Checks current state or history before making changes \\
M35 & Ordered change sequence & Checks, stages, then commits a change with the given description \\
M36 & Create then switch & Creates a new context, then switches to it as a separate step \\
M37 & Exact names & Resolves names exactly as given \\
M38 & Required format & Uses the value format the tool requires \\
M39 & Range query & Uses one range query in place of a call per item \\
M40 & Specialised lookup & Uses the dedicated lookup tool for an identifier system \\
M41 & Right reader for the format & Uses the reader made for the source's format, not a generic one \\
M42 & Fix after validation error & Retries with parameters corrected from the error \\
M43 & Nearest valid input & Tries the nearest valid input when the given one has no data \\
M44 & Different tool after empty result & Moves to a more direct query or a known identifier \\
M45 & Read before editing & Reads a source before making a precise edit \\
M46 & Count before changing & Counts occurrences, then makes the modification \\
M47 & Read all before comparing & Reads every source before a comparison or join \\
M48 & Filter before random pick & Filters first, then passes only that subset on \\
M49 & Computed values are own work & Computed figures come from its own calculation on retrieved data \\
M50 & Units in answer & States units with quantities \\
M51 & Show the calculation & Gives the formula or breakdown for computed results \\
M52 & Format for the target & Formats structured data for the target's type \\
M53 & Current time from a service & Retrieves the date or time; doesn't assume it \\
M54 & One call per target & Converts separately for each target \\
M55 & All calls then one answer & Makes every needed call, then gives a combined answer \\
M56 & One request if the split fails & Treats the whole task as one request when the split gives nothing usable \\
M57 & Search in the tool's form & Uses the form the search tool requires and never repeats a search unchanged \\
M58 & Call values from the task or results & Values in a call come from the task or a tool result \\
M59 & Empty is not success & Treats an empty result, or one that fails for every item, as not a success \\
M60 & Drop a dead source & Does not use again a source that fails to connect or lacks the tool \\
M61 & Budget in calls & Counts the planner budget in calls per action, not seconds \\
\end{longtable}
\end{center}

\subsection{Telecom, 64 rules}

\begin{center}\footnotesize\setlength{\tabcolsep}{4pt}
\begin{longtable}{@{}lp{0.6\linewidth}@{}}
\toprule
\textbf{Rule} & \textbf{What the expert does} \\
\midrule
\endhead
\bottomrule
\endfoot
Identify first & Looks up the user before anything else \\
Target details first & Reads the target record's details before target-specific work \\
Confirm the fix at the end & Runs the task's success check to confirm the fix \\
Grant each permission & Grants each required permission separately \\
Clear the blocking state first & Clears a state that blocks everything else before other troubleshooting \\
Check detailed status & Reads detailed state on connectivity problems \\
Overview early & Checks a quick overview early \\
Restart after a settings reset & Always follows a settings reset with a restart \\
Enable on the account first & Enables a feature on the account before the user toggles it on the device \\
Check app permissions & Verifies the relevant app's permissions \\
Fix the mode setting & Sets a restrictive mode to the preferred one \\
Verify after changes & Runs a check to verify service after changes \\
Check configuration & Verifies the configuration values \\
Reset bad configuration & Resets and restarts when a configuration value is wrong \\
Verify usage before adding & Checks usage before adding an allowance \\
Enable the disabled service & Turns a service on when it is disabled \\
Reseat a missing component & Asks the user to reseat a component that is not detected \\
Find the right record & Reads all records to match the user's identifier \\
Toggle on the device after the account & Toggles the feature on the device after enabling it on the account \\
Interfering feature off & Turns off an interfering feature during troubleshooting \\
Check component status & Verifies the component is active on service issues \\
Read plan details & When limits or prices matter \\
Check the problem early & Confirms the reported problem before attempting fixes \\
Check for interference & Looks for interfering features during troubleshooting \\
Restart after resuming & Asks for a restart right after resuming service \\
Disconnect a slow intermediary & When an intermediary is connected and slow \\
Limiter off & Turns a limiter off when it is on \\
Maximum allowed amount & Adds the maximum amount allowed \\
Bills before resuming & Checks for overdue bills on a suspended account \\
Diagnose before changing & Runs user-side checks before account changes \\
Mode setting early & Checks the mode setting early in troubleshooting \\
Read before changing a record & Never changes settings without reading them first \\
Customer acts on their device & Never calls user-side tools directly \\
Verify payment & Confirms the bill shows as paid before proceeding \\
Payment request first & Sends it before asking the user to pay \\
Both permissions & Grants every missing permission if any is missing \\
Toggle if still off & When the device shows the feature off after enabling it \\
Locked component to a human & Transfers at once when a component needs an unlock code \\
Restart after setting changes & After configuration or mode changes \\
Underlying service first & Fixes the underlying service first when both are reported \\
Reseat after restart fails & On no-signal problems \\
Check installed apps & When the app's name is uncertain \\
Blocking state before toggles & Never toggles services while a blocking state is on \\
Ended contract to a human & Transfers without trying to resume service \\
Cycle the blocking state & Toggles it on then off before restarting, on no signal \\
One step at a time & Executes a procedure written as a list of steps one step at a time \\
Argument names the target & An identifying argument names the target, not the change to make \\
Refusal cites the rule & A refusal cites a value the rule's text mentions \\
Quantities from records & Every quantity is user-stated or computed from records \\
Target in the named state & When the request names a state, the target chosen is in that state \\
Escalate after a failed attempt & Escalates only after the action was attempted and refused or failed \\
Dates computed in code & Computes time conditions in code against the date the rules give \\
Fill from the record & Takes a value the user did not give from the record and confirms it \\
User's fallback & Applies the user's stated fallback when the request cannot be met \\
Permitted equivalent & Uses an equivalent action a rule permits when the rule blocks the requested one \\
Refuse only on a rule & Refuses only when the rule check produced the refusal \\
Explicit agreement & Agreement is explicit, covers one state change, and is not inferred from a choice among options or a qualified yes \\
Confirm once & Does not confirm an agreed state change a second time \\
Agreed change first & Carries out an agreed state change before anything else \\
Irreversible last & Does irreversible actions last, after telling the user they are final \\
Correction everywhere & Applies a correction everywhere the corrected value appears \\
Drop a no-op & Drops an action that would change nothing \\
Checks stay internal & Does not surface internal checks to the user as questions \\
No early escalation & Does not escalate while the user is still willing to continue \\
\end{longtable}
\end{center}

\section{The ReAct planner loop}
\label{app:react}

Each customer turn runs the loop below until the student emits an action.

\begin{center}\footnotesize\setlength{\tabcolsep}{5pt}
\begin{tabular}{@{}lp{0.78\linewidth}@{}}
\toprule
\textbf{Step} & \textbf{What happens} \\
\midrule
Plan    & Granite-4.2-8B reads the instructions, the tools, and the conversation and returns the situation, the applicable rules, and the next step. \\
Act     & The student continues from that plan, placed where its own reasoning would go, and emits one action. \\
Reflect & One Granite critic per aspect reviews the action; on a revise verdict the student retries with the critique. After the last revision round, the last action is returned. \\
\bottomrule
\end{tabular}
\end{center}

\begin{center}\footnotesize\setlength{\tabcolsep}{5pt}
\begin{tabular}{@{}ll@{}}
\toprule
\textbf{Setting} & \textbf{Value} \\
\midrule
Granite input-token budget per task & 450,000 (estimated as characters$/4.2 + 60$ per call) \\
critics & policy, grounding \\
maximum revision rounds & 2 \\
critic share of the budget & 0.5 \\
planning & on \\
tool-description cap & 500 characters \\
planner calls per customer turn & 60 \\
student \texttt{max\_tokens} & 3,000 \\
\bottomrule
\end{tabular}
\end{center}

\section{Sample-and-judge: the judge prompt}
\label{app:judge}

The judge is Granite-4.2-8B. Identical candidate next steps are pooled, and
each option is shown with the number of the five samples that proposed it.

\paragraph{System.}
\begin{lstlisting}[style=prompt]
You are the selection module of a tool-using assistant. The assistant produced several candidate next steps (OPTIONS), each with the number of independent samples that proposed it. Pick the ONE option that best follows the INSTRUCTIONS and is best supported by the CONVERSATION SO FAR and the tool results. When options are equally good, prefer the one proposed by more samples. Answer with one JSON object: {"choice": <option number>, "reason": str}
\end{lstlisting}

\paragraph{User.}
\begin{lstlisting}[style=prompt]
INSTRUCTIONS:
{system or '(none)'}
TOOLS:
{render_tools(tools)}
CONVERSATION SO FAR:
{render_conv(conv)}
OPTIONS:
1. [{votes} of {n} samples] {show(option)[:3000]}
2. [...]
\end{lstlisting}

\section{GEPA baseline: optimising the student's prompt}
\label{app:gepa-backbone}

The prompt-optimisation arm of \S\ref{sec:baselines}: the student's own system
prompt is optimised with GEPA~\citep{agrawal2026gepa} and the student is then
run as a native tool-calling agent with no program. The budget was 10
iterations; the held-aside score was plateauing by then, so more
iterations were not run.

\begin{center}\footnotesize\setlength{\tabcolsep}{5pt}
\begin{tabular}{@{}ll@{}}
\toprule
\textbf{Setting} & \textbf{Value} \\
\midrule
data & the training and held-aside splits of \S\ref{sec:data}, the same as the SFT arm \\
budget & 10 iterations, minibatch 16, 8 workers \\
reflection model & \texttt{openai/gpt-4.1} \\
reward & outcome only: exact call-set match; no process reward \\
parent selection & Pareto front over per-instance scores; $\max(\mathrm{pareto}(\mathrm{pool}), \mathrm{key}=\mathrm{mean})$ \\
accept rule & child's minibatch mean exceeds parent's; accepted children join the pool \\
selection & best score on the held-aside split \\
\bottomrule
\end{tabular}
\end{center}

\paragraph{Outcome per student.}

\begin{center}\footnotesize\setlength{\tabcolsep}{6pt}
\begin{tabular}{@{}lrr@{}}
\toprule
\textbf{Student} & \textbf{Best train} & \textbf{Best val} \\
\midrule
Nemotron 3 Super       & 0.838 & 0.950 \\
Nemotron 3 Nano        & 0.818 & 0.906 \\
Nemotron 3.5 Lightning & 0.800 & 0.805 \\
\bottomrule
\end{tabular}
\end{center}

\end{document}